\pdfoutput=1

\documentclass[11pt]{article}

\usepackage[final]{acl}

\usepackage{times}
\usepackage{latexsym}

\usepackage[T1]{fontenc}

\usepackage[utf8]{inputenc}

\usepackage{microtype}

\usepackage{inconsolata}

\usepackage{hyperref}
\usepackage{url}
\usepackage{multirow}
\usepackage{xcolor}
\usepackage{latexsym}
\usepackage{soul}
\usepackage{amsfonts}
\usepackage{amsmath}
\usepackage{amssymb}
\usepackage{nicefrac}
\usepackage{booktabs} 
\usepackage{makecell} 
\usepackage{graphicx}
\usepackage[inline]{enumitem}
\usepackage{subcaption}
\usepackage{caption}
\usepackage{xargs} 

\title{On the Effects of Fine-tuning Language Models for Text-Based Reinforcement Learning}

\author{Maur\'{i}cio Gruppi \\
  Villanova University\\
  \texttt{mgouveag@villanova.edu} \\\And
  Soham Dan \\
  IBM Research \\
  \texttt{soham.dan@ibm.com} \\\AND
  Keerthiram Murugesan \\
  IBM Research \\
  \texttt{keerthiram.murugesan@ibm.com}\\\And
  Subhajit Chaudhury \\
  IBM Research \\
  \texttt{subhajit@ibm.com}
  }

\begin{document}
\maketitle

\begin{abstract}
Text-based reinforcement learning involves an agent interacting with a fictional environment using observed text and admissible actions in natural language to complete a task.
Previous works have shown that agents can succeed in text-based interactive environments even in the complete absence of semantic understanding or other linguistic capabilities.
The success of these agents in playing such games suggests that semantic understanding may not be important for the task.
This raises an important question about the benefits of LMs in guiding the agents through the game states.
In this work, we show that rich semantic understanding leads to efficient training of text-based RL agents. Moreover, we describe the occurrence of semantic degeneration as a consequence of inappropriate fine-tuning of language models in text-based reinforcement learning (TBRL).
Specifically, we describe the shift in the semantic representation of words in the LM, as well as how it affects the performance of the agent in tasks that are semantically similar to the training games. We believe these results may help develop better strategies to fine-tune agents in text-based RL scenarios.

\end{abstract}
\section{Introduction}
\label{sec:intro}

Text-based games (TBGs) are a form of interactive fiction where players use textual information to manipulate the environment.
Since information in these games is shared as text, a successful player must hold a certain degree of natural language understanding (NLU).
TBGs have surfaced as important testbeds for studying the linguistic potential of reinforcement learning agents along with partial observability and action generation.
TBGs can be modeled as partially observable Markov decision processes (POMDP) defined by the tuple $\langle S, A, O, T, E, R \rangle$, where $S$ is the set of states, $A$ the set of actions, $O$ the observation space, $T$ the set of state transition probabilities, $E$ is the conditional observation emission probabilities, and $R: S \times A \rightarrow \mathbb{R}$ the reward function. 
The goal of a TBG agent is to reach the end of the game by interacting with the environment through text, while maximizing the final score.

\begin{figure}
    \centering
    \includegraphics[width=\linewidth]
    {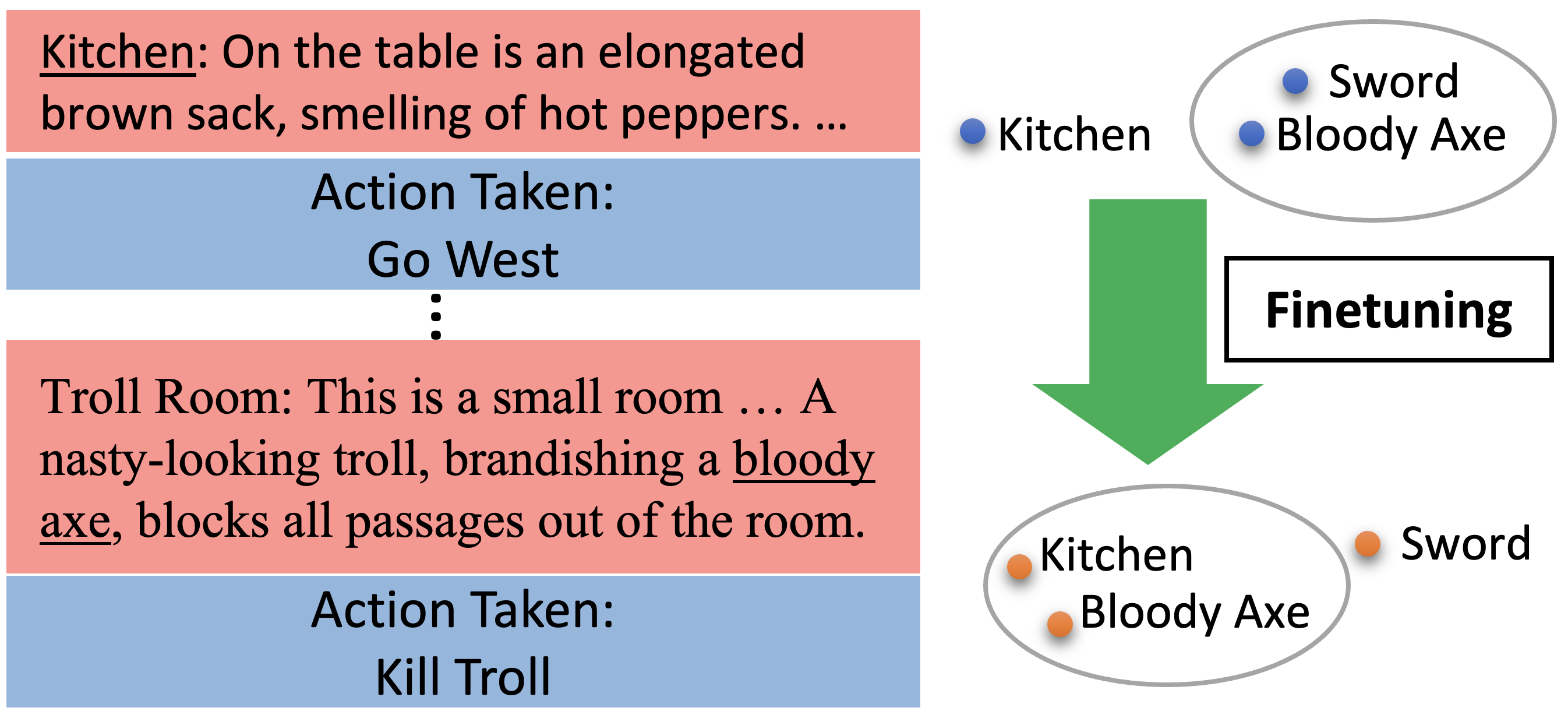}
    \caption{Semantic degeneration of the terms \textit{kitchen} and \textit{bloody axe} in \textit{Zork 1}.}
    \label{fig:overview}
\end{figure}

In TBGs, observations and actions are presented in the form of unstructured text, therefore, they must be encoded before being passed onto the RL network.
Recent works in text-based RL adopt a strategy where such encoding is learned from the game, typically by fine-tuning a language model, such as embeddings or transformers, using the rewards values from the training \cite{yao2020keep,wang2022behavior}. 
We hypothesize that this approach may cause the language model to overfit the training games, leading to the degeneration of the semantic relationships learned during pretraining, and, subsequently, negatively impacting the agent's training efficiency and transfer learning capacity.
We conduct experiments in two distinct TBG domains: (1) TextWorld Commonsense (\texttt{TWC}) \citep{murugesan2021text}, and (2) Jericho \citep{hausknecht19}.
The former provides a number of games where the goal is to perform house cleaning tasks such as taking objects from a location and placing them in their appropriate places, using commonsense knowledge. 
The latter provides a library of classic text-adventure games, such as the \emph{Zork} (1977), each having its own unique objectives, characters, and events. Unlike \texttt{TWC} games, Jericho games may not let the player know a priori what the final goal is. Instead, the player is expected to explore the game to learn the story and complete the tasks one-by-one.
In both domains, the agent selects an action from a given list.

Under this framework, we address the following research questions:
\begin{enumerate}
    \item Does fine-tuning the language model to the RL rewards improve the training efficiency in comparison to fixed pre-trained LMs?
    \item Does fine-tuning LMs make agents robust to tasks containing out-of-training vocabulary?
\end{enumerate}

Our goal is to evaluate what are the implications, pros, and cons of fine-tuning LMs to the RL tasks.
Our results indicate fine-tuning LMs to rewards leads to a decrease in the agent's performance and hinders its ability to play versions of the training games where the observations and actions are slightly reworded, such as through paraphrasing or lexical substitution (synonyms).
In comparison to fixed pre-trained LMs, these fine-tuned agents under-performed in training and in test settings.
We refer to this process as \textbf{semantic degeneration}, because it leads to loss of relevant semantic information, in the LM, that would be crucial to produce generalizable representation.
For instance, by learning that the terms ``bloody axe'' and ``kitchen'' are related to each other in the game \textit{Zork 1}, the agent overfits to this setting and, in turn, loses relevant information about ``kitchen'' and ``bloody axe'' that could be important to other games.
In NLP, semantic generation might be an expected consequence of fine-tuning \cite{mosbach2020stability}, however, the vast majority of text-based RL agents employ LMs that are fully fine-tuned to the game's semantics.





\section{Background}
\label{sec:method}

\paragraph{Model and Architecture}
\label{sec:model_and_architecture}

The general architecture of the agents in this work consist of a state encoder akin to the DRRN \citep{he2015deep} with an actor-critic policy learning \citep{wang2016sample} and experience replay.
The main components of the agent's network are (1) a text encoder, (2) a state-action encoder, and (3) an action scorer.
The text encoder module is a language model that converts an observation $o \in O$ and action $a \in A$ from text form to fixed length vectors $f(o)$ and $f(a)$.
The state-action encoder consists of a GRU \cite{dey2017gate} that takes as input the encoded state and actions, and predicts the Q-values for each pair: $Q_\phi(o, a) = g(f(o), f(a))$ given parameters $\phi$.
The action predictor is a linear layer that outputs the probabilities based on the Q-values from the previous layer. 
The chosen action is drawn following the computed probability distribution.
The agent is trained by minimizing the temporal differences (TD) loss: $\mathcal{L}_{TD}=(r + \gamma \max_{a' \in A}Q_\phi (o', a') - Q_\phi(o, a))^2$ where $o'$ and $a'$ are the next observation and next actions sampled from a replay memory, $\gamma$ is the reward discount factor.

\paragraph{Text Encoders}

We used three distinct types of encoders in this study:

\vspace{-1em}
\begin{itemize}
    \item {\textit{Hash} -  does not capture semantic information from the text. It utilizes a hash function to reduce the observation to a random vector. Similarly to \citet{yao2021reading}}.  \vspace{-0.75em}  
    \item {\textit{Word Embedding} - pre-trained static GloVe embeddings \cite{pennington2014glove} and a GRU to encode the sequences of tokens.} \vspace{-0.75em}
    \item {\textit{Transformers} - pre-trained LMs to encode observations \citep{devlin2018bert, liu2019roberta}}. \vspace{-0.75em}
\end{itemize}
These encoders are often the top performer \cite{murugesan2021efficient,ammanabrolu2020graph,wang2022scienceworld,atzeni2021case,yao2020keep,tuyls2021multi} in benchmark environments for text-based reinforcement learning such as Textworld \cite{cote2018textworld}, Jericho \cite{hausknecht19}, Scienceworld \cite{wang2022scienceworld}, etc.

\section{Results}
\label{sec:experiments}

We now present our main results.
In the TWC environment, agents are trained for 100 episodes, with a maximum of 50 steps per episode (repeated over 5 runs).
In the Jericho environment, agents were trained over 100000 steps with no limit to the number of episodes (repeated over 3 runs). 
These settings were chosen following previous work reference in this manuscript, such as \citet{yao2021reading} and \citet{murugesan2021efficient}.

We deploy agents of the same architecture as described in Section \ref{sec:model_and_architecture}, the only exception being that the input encoder used by them is different.
The encoders are the \emph{Hash} encoder, which produces semantic-less vectors, the Word \emph{Embedding} which uses pre-trained GloVe embeddings, and the transformer LMs Albert \cite{lan2019albert} and RoBERTa \cite{liu2019roberta}.
The transformer encoders are used in two variations: \textit{Fixed}, where the LMs weights are frozen; and \textit{Fine-tuned (FT)}, where the LMs weights are updated according to the rewards.
This allows us to compare the performance of the typical text-based RL fine-tuning approach to unconventional ones.

\subsection{Semantic Information from Pre-Training Improves the Overall RL Performance}
We evaluate the use of different LMs to encode the observations and actions into fixed-length vectors.
We begin our analysis with the weights of the language model-based encoders fixed, i.e., only the RL network parameters $\phi$ are updated.

\begin{figure}[t]
    \centering
        \begin{subfigure}{0.4\textwidth}
            \includegraphics[width=\textwidth]{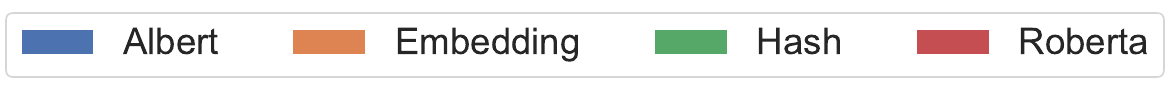}
        \end{subfigure}
        
        \begin{subfigure}{0.23\textwidth}
            \includegraphics[width=\textwidth]{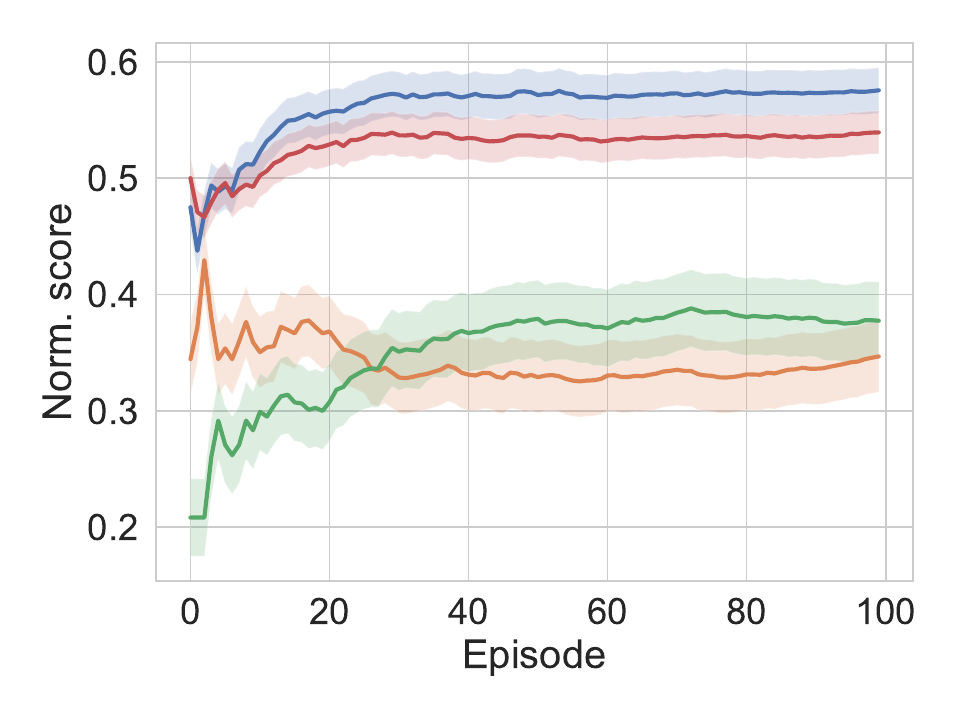}
        \end{subfigure}~
    \begin{subfigure}{0.23\textwidth}
        \centering
        \includegraphics[width=\textwidth]{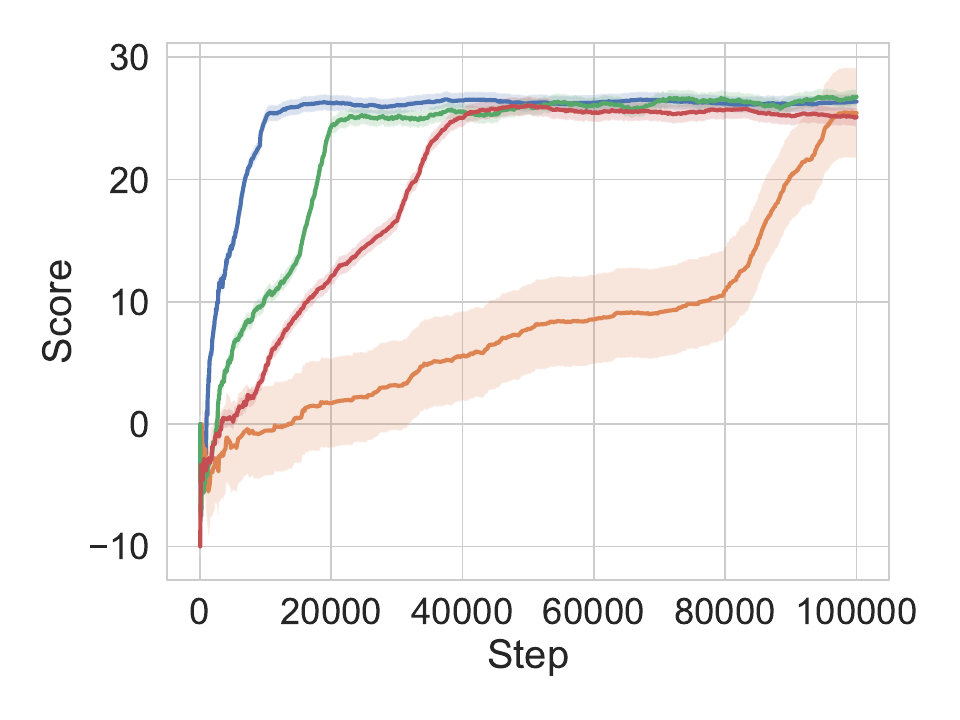}
    \end{subfigure}
    \caption{Training performance comparing LM-based encoding models and hash/word embedding-based models. (left) shows the normalized scores 
    for TWC games and (right) shows the game score achieved in training across 100k steps in \emph{Zork 1}. Shaded area corresponds to one standard deviation.}
    \label{fig:model-comparison}
\end{figure}

\vspace{-0.1cm}
\paragraph{The rich semantic information of LMs accelerates training:}
The results from these experiments show that even an agent without semantic information can properly learn to play the games.
However, an agent leveraging the semantic representations from language models are able to: (1) converge more quickly, in training, to a stable score than hash and simple, as shown in Figure \ref{fig:model-comparison}; (2) handle out-of-training vocabulary, Table \ref{tab:twc-medium-scores} shows the performance of the models under two settings: games using an in-training vocabulary (ID) and games using an out-of-training vocabulary (OOD). 
\begin{table}[t]
    \centering
    \small{
    \begin{tabular}{l|r|r}
\textbf{Model}            & \multicolumn{1}{c}{\textbf{ID}} & \multicolumn{1}{c}{\textbf{OOD}}  \\
    \toprule
    Hash & $0.58 \pm 0.06$ &  $0.15 \pm 0.03 $\\
    Embedding & $0.58 \pm 0.08$ & $0.43 \pm 0.07 $ \\
    Albert \tiny{\cite{lan2019albert}}* & $0.66 \pm 0.05$ & $\textbf{0.65} \pm 0.05 $ \\
    RoBERTa \tiny{\cite{liu2019roberta}}* & $\textbf{0.70} \pm 0.05$ & $0.53 \pm 0.06 $ \\
    \end{tabular}
    }
    \caption{Normalized scores for the in-distribution vocabulary (ID) and out-of-distribution vocabulary (OOD) game sets in TWC's Medium difficulty games. (*) Indicates fixed language models.}
    \label{tab:twc-medium-scores}
\end{table}
These results show that the fixed transformer LMs outperform the Hash and Embedding models in both vocabulary distributions, highlighting the importance of keeping the semantic information from pre-training intact.

\subsection{Semantic Degeneration Hurts Learning}

\begin{figure}[t]
    \centering

    \begin{subfigure}{0.4\textwidth}
    \centering
        \includegraphics[width=\textwidth]{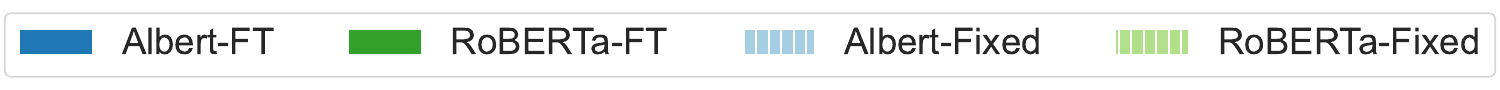}
    \end{subfigure}
    
    \begin{subfigure}{0.225\textwidth}
        \includegraphics[width=\textwidth]{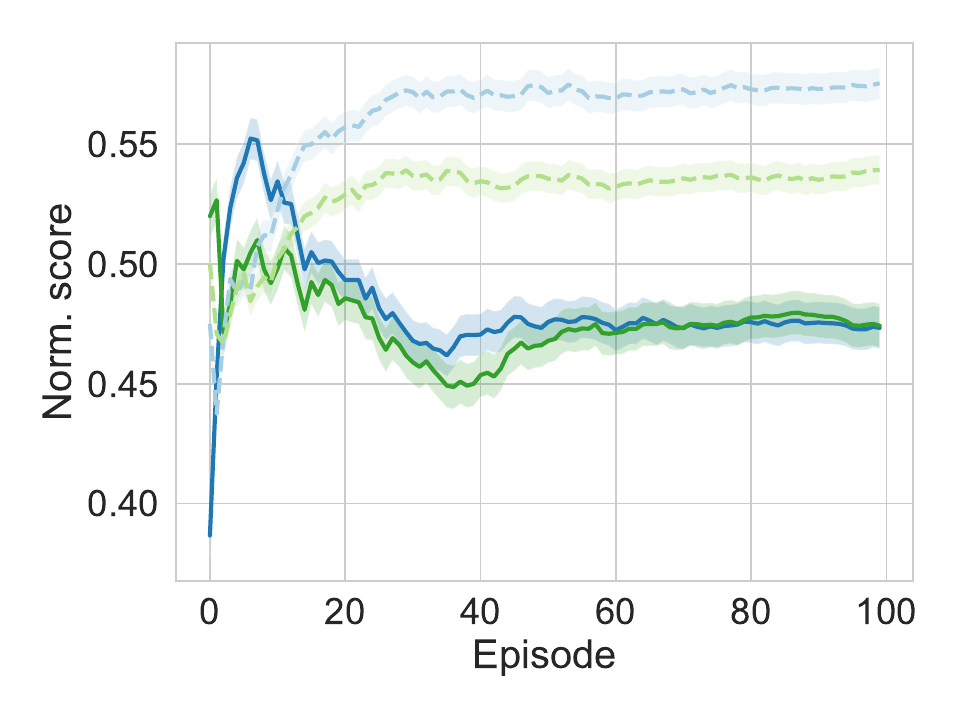}
    \end{subfigure}~
    \begin{subfigure}{0.225\textwidth}
        \includegraphics[width=\textwidth]{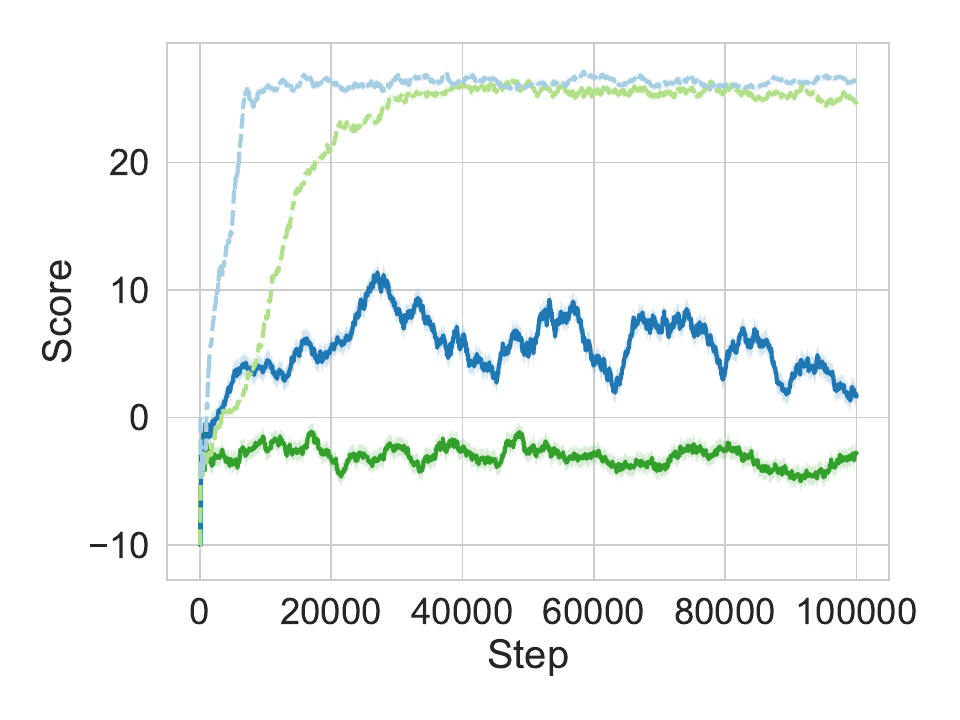}
    \end{subfigure}
    
    \caption{Training curves of fixed/fine-tuned LMs on (left) \texttt{TWC} medium difficulty games and (right) \textit{Zork 1}. Due to semantic degeneration, the fine-tuned models do not exhibit an increasing score converging to a maximum value. Shaded areas denote one standard deviation.}
    \label{fig:tuning_transformers}
\end{figure}

In this experiment, we address the first proposed research question: ``does fine-tuning the LM to the RL rewards improve the training efficiency in comparison to fixed pre-trained LMs?''
To that end, we trained the Fixed and Fine-tuned variations of Albert and RoBERTa encoders on the same games, and compared their scores during training. Figure \ref{fig:tuning_transformers} shows the outcome of the experiment.
The findings suggest that traditional text-based RL approach of fine-tuning the LMs lead to substantially lower training scores, which are due to semantic degeneration. 
That is, \textbf{semantic degeneration leads to ineffective training of the RL agents}, whereas the fixed models converge to a higher score after a relatively small number episodes/steps.

Semantic degeneration arises from fine-tuning the LMs to the training rewards. 
The LM ``forgets'' its semantic associations it had learning during its pre-training, such as the masked token prediction in the case of transformers. This ``forgetting'' originates from overfitting the model's weights to the games' word distributions.
The biggest problem arises from the fact that the RL network receives the encoded vectors from the LM and updates its weights based on such initial representations. However, since the LMs are fine-tuned, the encoding will change between each episode, causing the RL network to receive a different encoding for the same observation as the training goes on.

\begin{figure}[t]
    \centering
        \includegraphics[width=0.235\textwidth]{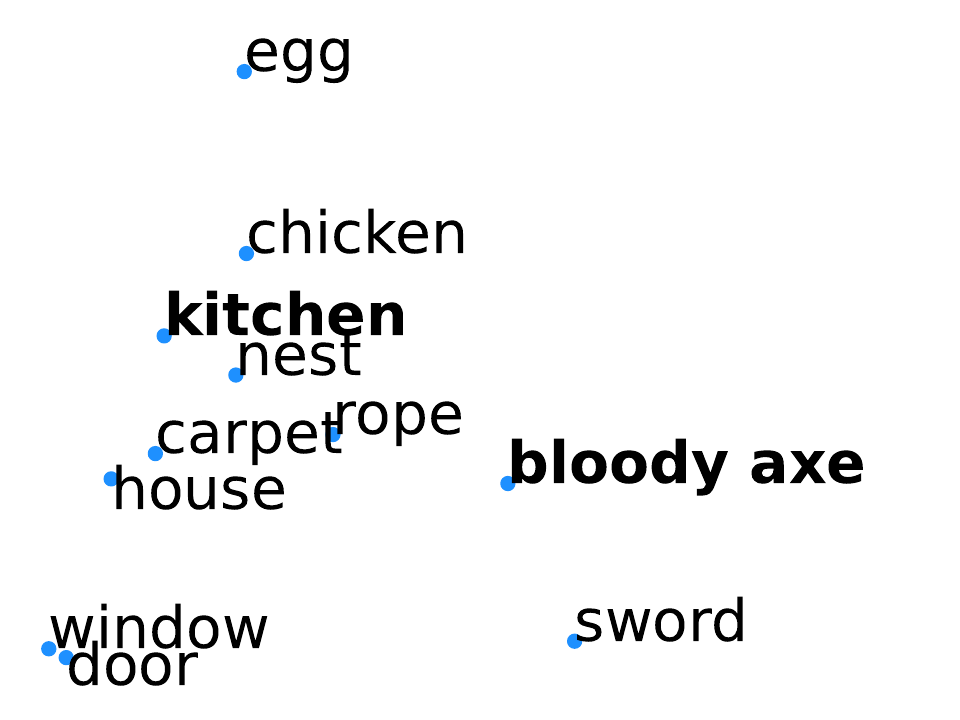}
        \includegraphics[width=0.235\textwidth]{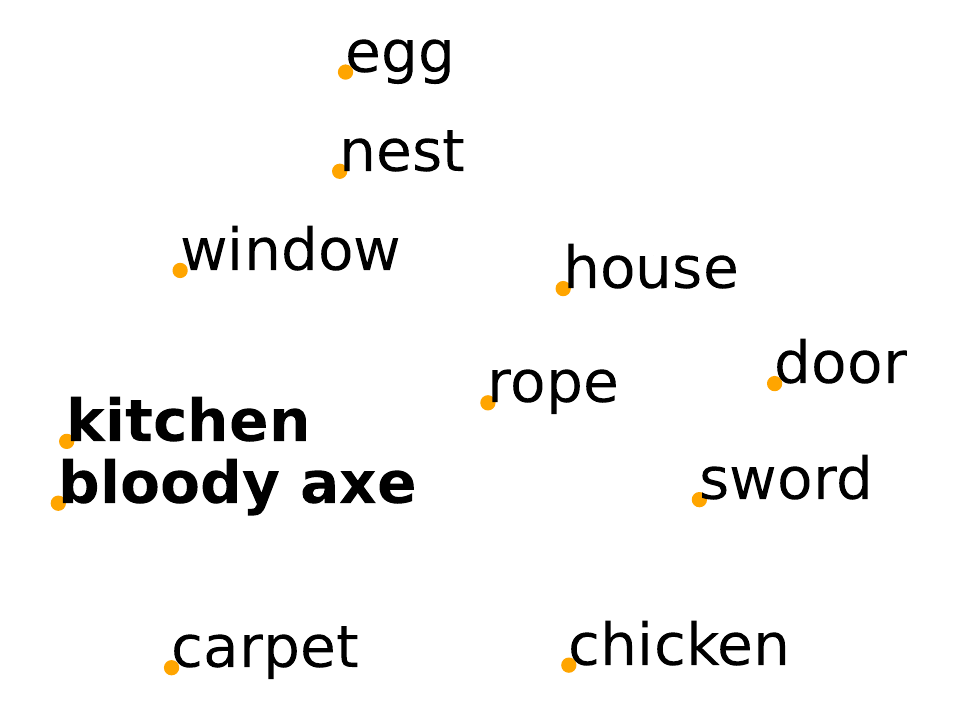}
    \caption{{Shift caused by the semantic degeneration to the contextual word vectors in the RoBERTa model fine-tuned to \emph{Zork 1}: (a) shows the word embeddings from the pre-trained model, (b) shows the word embeddings after fine-tuning to \emph{Zork 1}. The bold words denote the case where the term ``bloody axe'' shifts towards the word ``kitchen'' as a result of them co-occurring in a positively rewarded state.}}
    \label{fig:word-shift}
\end{figure}

A comparison of pre-trained and semantically degenerated word vectors is seen in Figure \ref{fig:word-shift}. 
A 2D TSNE plot of pre-trained word vectors from a RoBERTa model is seen in Figure \ref{fig:word-shift}a; Figure \ref{fig:word-shift}b shows the plot of the word vectors after fine-tuning the LM to the game \textit{Zork 1}.
Notice the shift of the term ``bloody axe'' towards the term ``kitchen'' from (a) to (b). The shift happens because both terms appear in a sequence early on in the game, therefore, the association between their vectors becomes stronger as the LM is fine-tuned. Moreover, the terms ``egg'' and ``nest'' shift away from ``chicken''. The first two terms are also employed in the game in a sequence where the agent receives a positive reward, whereas the last is never used in the game.
Despite being related \textit{in-game}, these terms should have their semantic relationships preserved, which is possible by utilizing fixed LMs. 


\subsection{Agents with fine-tuned LMs are less robust to language change}
\label{sec:perturbations}

We address the second research question: ``does fine-tuning LMs make agents robust to tasks containing out-of-training vocabulary?''.
To test the robustness of each model, we first train each agent on a particular game. 
Then, we evaluate the agents by having them play games where the observations are transformed in one of the following ways: Paraphrasing, we run the observations through a paraphrasing model to rephrase the descriptions (using a Bart-based paraphrase \cite{lewis2019bart}); Lexical Substitution, we replace words in the observations using synonyms and hypernyms from WordNet \cite{fellbaum2010wordnet}.
By playing these versions of the games, agents have to perform the same task as seen in training, but with reworded or slightly modified observations.

\begin{figure}[ht]
    \centering
    \begin{subfigure}{0.235\textwidth}
        \includegraphics[width=\textwidth]{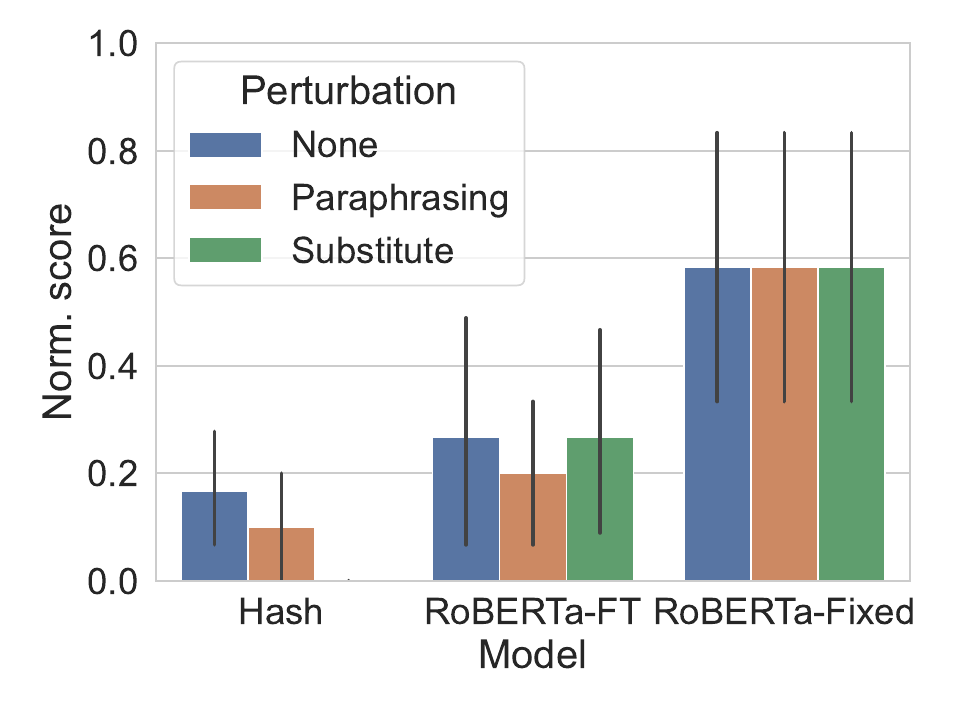}
    \end{subfigure}
    \begin{subfigure}{0.235\textwidth}
        \centering
        \includegraphics[width=\textwidth]{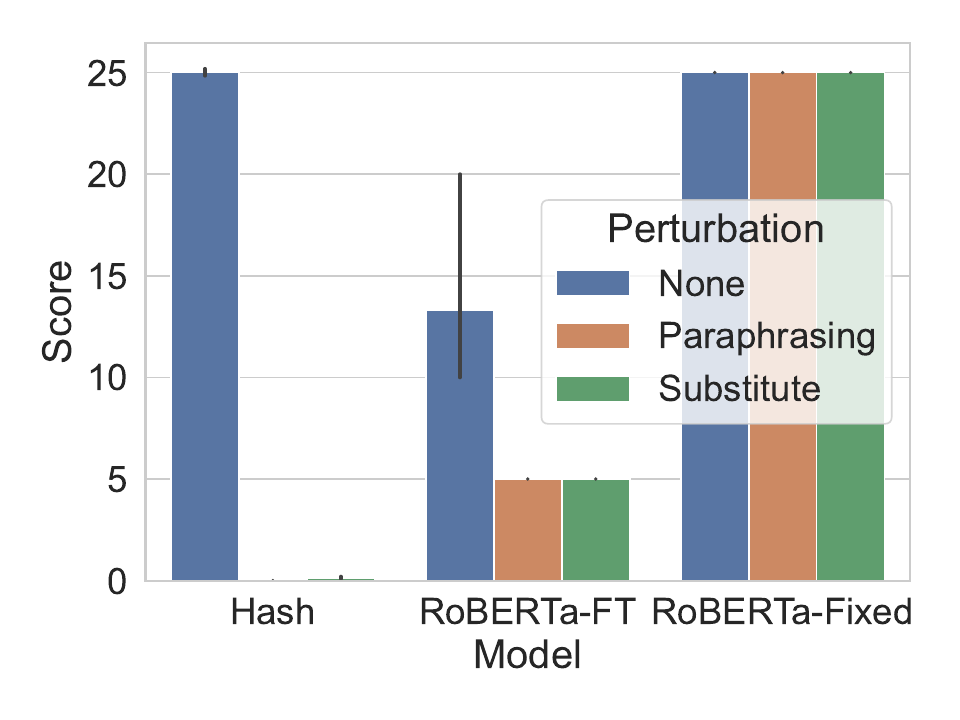}
    \end{subfigure}

    \caption{Evaluation of a RoBERTa agent on original (none), paraphrased, and lexical substitution observations on (left) TWC medium games and (right) \emph{Zork 1}. In both scenarios, fixed LMs exhibit strong robustness to the perturbations, scoring as much as in the games without perturbations.}
    \label{fig:perturbations}
\end{figure}

Figure \ref{fig:perturbations}a shows the fixed LM agent is robust to paraphrasing as it is able to maintain the original score even in the modified versions. 
This is due to the ability of LMs to handle such perturbations in text.
This evidence emphasizes the hypothesis that semantic understanding is important for generalization to words unseen in training.
Figure \ref{fig:perturbations}b shows the performance of the three agents in \emph{Zork 1}.
The fine-tuned agent exhibits a decline in performance while playing the paraphrased and lexical substitution games.
This is explained by the fact that the LM has been adjusted to the semantics of the original game, thus, tokens are no longer distributed according to semantic similarity.
The hash-based agent is unable to score in either of the modified games due to the lack of semantic information. 
The fixed agent, however, exhibits strong robustness to the perturbations.
This shows how semantic degeneration leads to decrease in performance in unseen or slightly different games.


\section{Conclusion}
In this paper, we have put forth a novel perspective over the occurrence of semantic degeneration at the intersection of LM fine-tuning and text RL. 
We have shown that semantic understanding brings benefits to the training of agents.
Moreover, despite being the typical approach to text-based RL, learning the semantics from the game may not be the optimal approach to training agents.
Our results corroborate the well known trends of trading-off general semantics for task-specific representations in NLP tasks; we shine light on how this affects agents in carrying out tasks that are semantically similar to the training ones.
Our results indicate that using meaningful semantic representations can be beneficial, and fine-tuning strategies may be developed to ensure prior semantic information is not lost by the model, while learning task-specific representations.

\section*{Limitations}
Our work focuses on popular TBG environments and also popular choices of LMs. In future work it would be interesting to study rarer TBG environments, potentially beyond English. In that context it would also be interesting to study multilingual LMs as the semantic representation for these games. Since we use LM representations for game playing, some of the limitations of these representations (like inability to distinguish between some related concepts, or certain biases), might carry over. Investigating these in detail is another interesting avenue to be explored.



{
\bibliography{tbg}
}

\appendix
\section{Appendix}

\subsection{TextWorld Commonsense}

This section contains information about the games (Table \ref{tab:twc-games}) in TextWorld Commonsense as well as an example of an observation and plausible actions (Figure \ref{fig:twc-example}).

The goal of TWC games are to complete a series of household tasks, such as ``picking up an apple and putting it in an appropriate location''. The agent is provided with the description of a scene and a list of plausible actions. They must then decide which action to be taken in the current game state. If the action performed is good, the agent is rewarded with points.

TWC games are split into easy, medium and hard difficulties. As the difficulty increases, the number of target objects and rooms to cleanup increases. Details can be seen in Table \ref{tab:twc-games}.

 \begin{table}[t]
     \centering
     \begin{tabular}{l|r|r|r}
          & \textbf{Objects} & \textbf{Targets} & \textbf{Rooms}  \\
          \toprule
          \textbf{Easy} & 1 & 1 & 1 \\
          \textbf{Medium} & 2--3 & 1--3 & 1 \\
          \textbf{Hard} & 6--7 & 5--7 & 1--2
     \end{tabular}
           \caption{No. of objects, target objects and rooms in \texttt{TWC} games per difficulty level.}
     \label{tab:twc-games}
 \end{table}

  \begin{figure*}[ht]
     \centering
     \fbox{
     \begin{minipage}{0.6\linewidth}
        \textbf{Observation}
        
        You've entered a kitchen.
        
        Look over there! A dishwasher.
        You can see a closed cutlery drawer.
        You see a ladderback chair.
        On the ladderback chair you can make out a dirty whisk.
     \end{minipage}
     }~
    \fbox{
     \begin{minipage}{0.3\linewidth}
     \textbf{Plausible Actions}
     \setlist{nosep}
     \begin{itemize}[leftmargin=*, label={}]
        \item Open dishwasher
        \item Open cutlery drawer
        \item Take dirty whisk from ladderback chair
     \end{itemize}
     
     \end{minipage}
    }
     \caption{Example of an observation from a TextWorld Commonsense game.}
     \label{fig:twc-example}
 \end{figure*}

\subsection{Model comparison in TWC}

\begin{figure*}[t]
    \centering
    
    \begin{subfigure}{0.9\textwidth}
        \includegraphics[width=\textwidth]{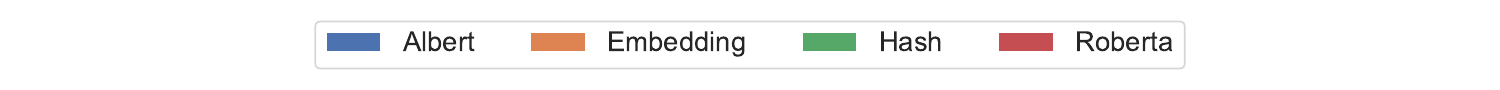}
    \end{subfigure}
    
    \begin{subfigure}{0.32\textwidth}
        \centering
        Easy
    \end{subfigure}~
    \begin{subfigure}{0.32\textwidth}
        \centering
        Medium
    \end{subfigure}~
    \begin{subfigure}{0.32\textwidth}
        \centering
        Hard
    \end{subfigure}~
    
    \begin{subfigure}{0.32\textwidth}
        \includegraphics[width=\textwidth]{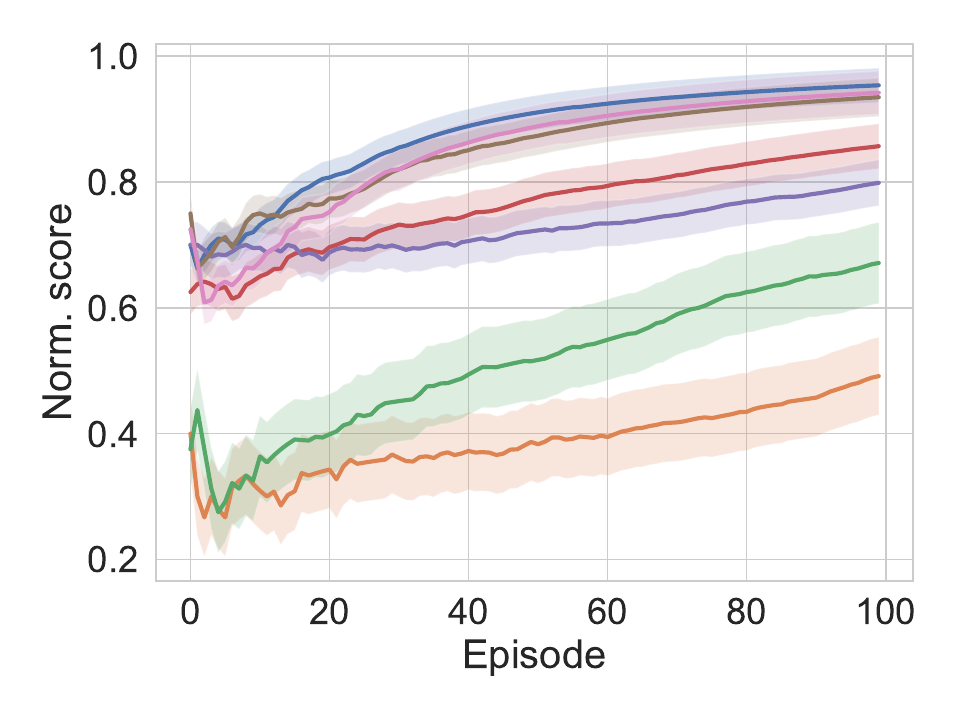}
        \caption{}
    \end{subfigure}~
    \begin{subfigure}{0.32\textwidth}
        \includegraphics[width=\textwidth]{img/model_comparison/model_comparison_medium_no_tuning_train_norm_scores.pdf}
        \caption{}
    \end{subfigure}~
    \begin{subfigure}{0.32\textwidth}
        \includegraphics[width=\textwidth]{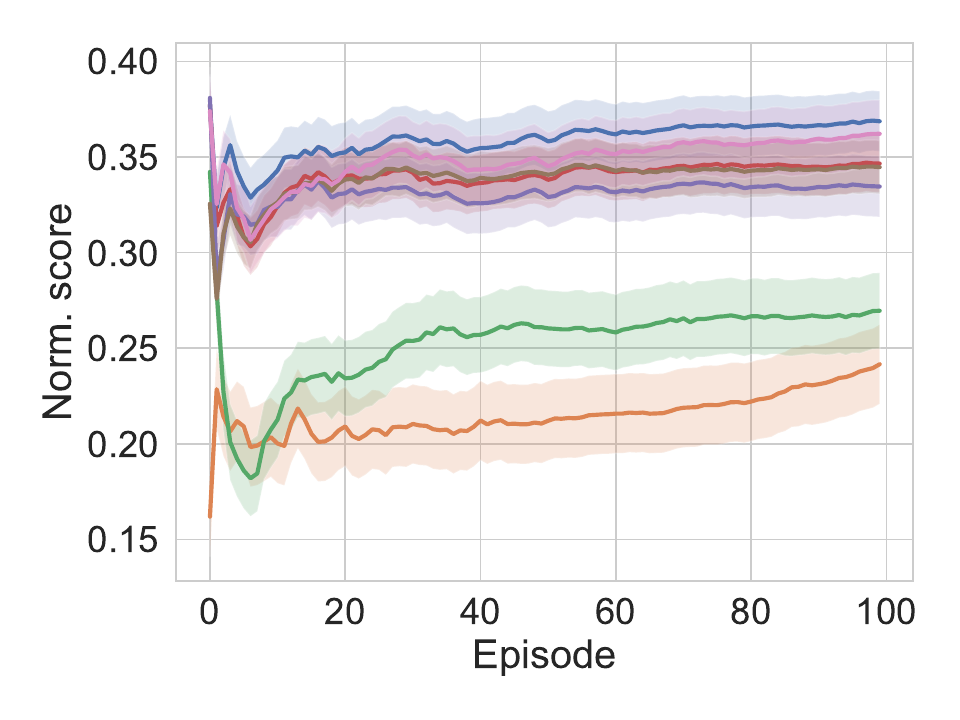}
        \caption{}
    \end{subfigure}
    
    \begin{subfigure}{0.32\textwidth}
        \includegraphics[width=\textwidth]{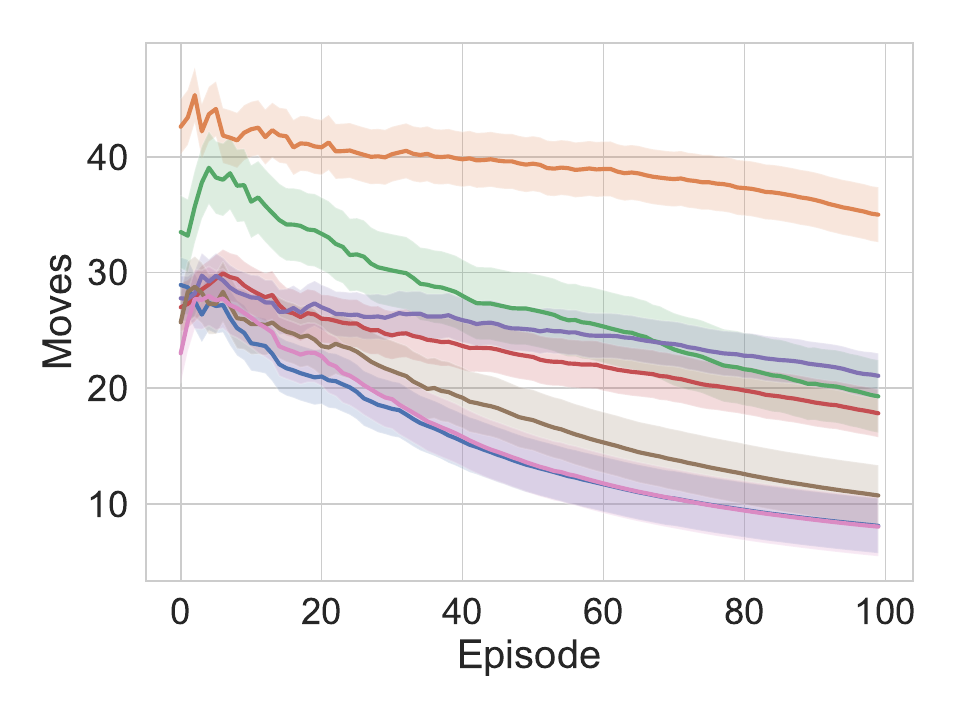}
        \caption{}
    \end{subfigure}~
    \begin{subfigure}{0.32\textwidth}
        \includegraphics[width=\textwidth]{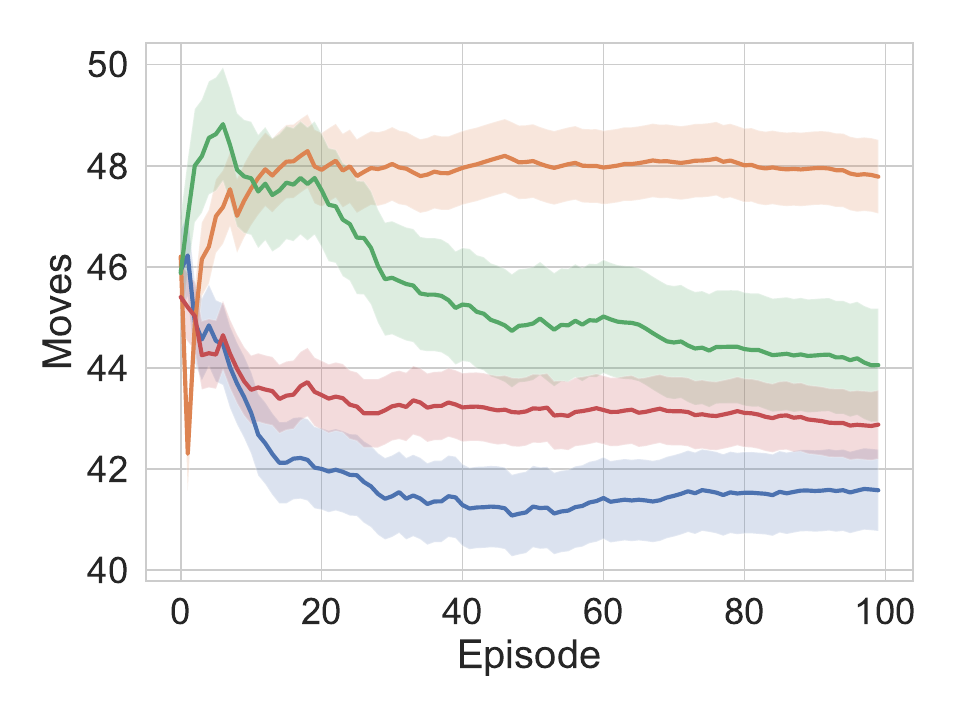}
        \caption{}
    \end{subfigure}~
    \begin{subfigure}{0.32\textwidth}
        \includegraphics[width=\textwidth]{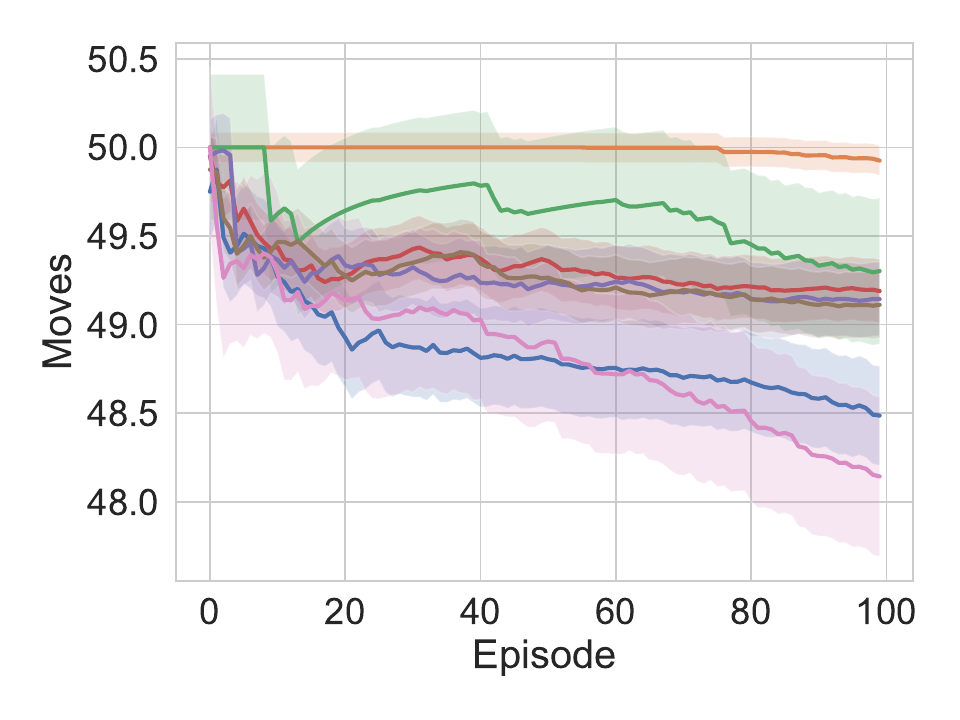}
        \caption{}
    \end{subfigure}

    \caption{Comparison of the performance across several language encoding models. Figures a, b, c show the normalized score for easy, medium and hard games, respectively. Figures d, e, f show the number of movements needed by the agent to complete the task (lower values are better). Shaded region corresponds one standard deviation.}
    \label{fig:model-comparison-full}
\end{figure*}

Figure \ref{fig:model-comparison-full} shows the comparison between all language models in all three difficulties of TWC in terms of normalized score and number of movements.

These results show how agents using fixed LMs converge earlier to a stable score (Figures \ref{fig:model-comparison-full} a, b, c) and to stable number of movements (Figures \ref{fig:model-comparison-full} d, e, f). Higher scores are better. Lower number of movements are better because it means the agent can complete the task while taking fewer actions, avoiding unnecessary moves.

\subsection{Complete Table of TWC Results}
\label{sec:twc-results-full}

Tables \ref{tab:twc-in-distro-full} and \ref{tab:twc-out-of-distro-full} show the results for all difficulties in TWC in the in-distribution set and out-of-distribution set.

We can see that fixed LMs consistently perform better when applied to both in-distribution and out-of-distribution tasks. This is due to the fact that they can keep rich semantic information and not suffering from semantic degeneration.

\subsection{Complete results for perturbation experiments in TWC}

\begin{figure*}[t]
    \centering

    \begin{subfigure}{0.32\textwidth}
        \centering
        \includegraphics[width=\textwidth]{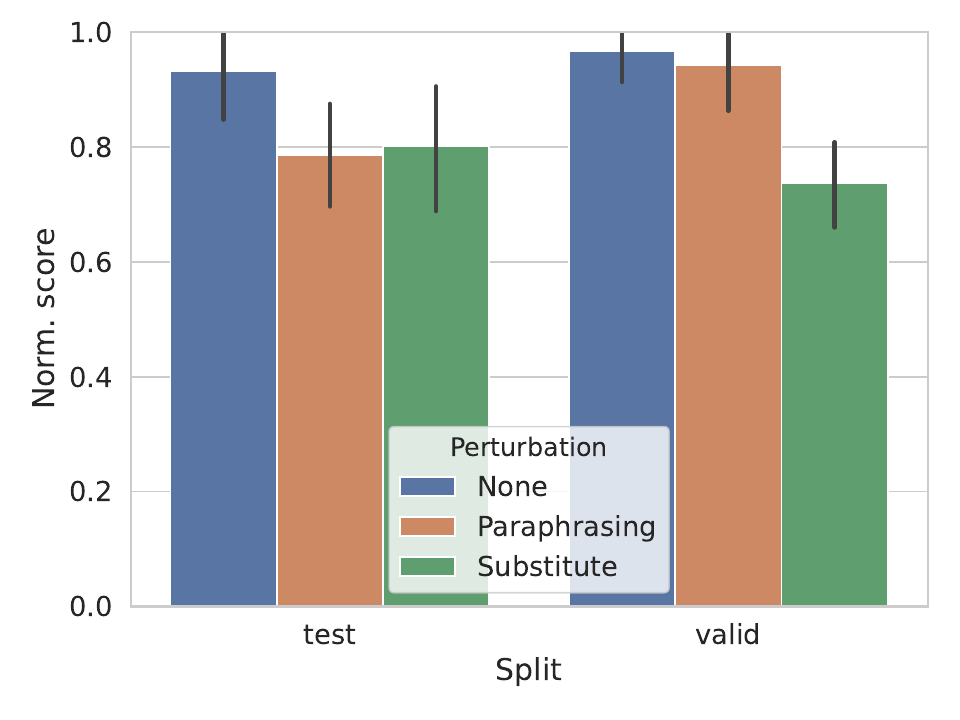}
        \caption{TWC Easy}
    \end{subfigure}
    ~
    \begin{subfigure}{0.32\textwidth}
        \includegraphics[width=\textwidth]{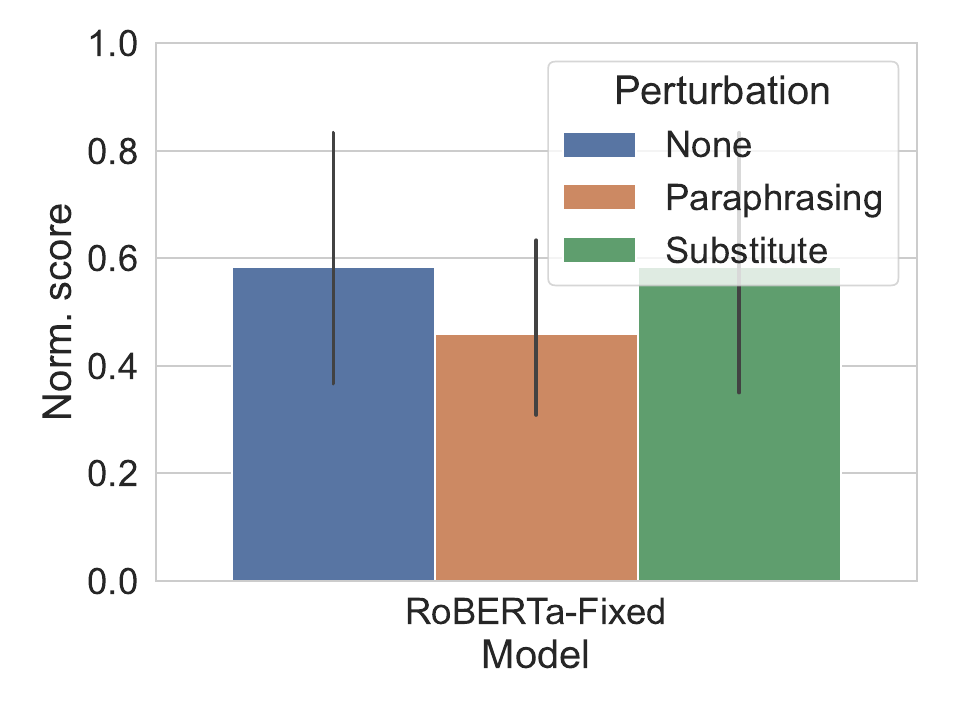}
        \caption{TWC Medium}
    \end{subfigure}
    \begin{subfigure}{0.32\textwidth}
        \includegraphics[width=\textwidth]{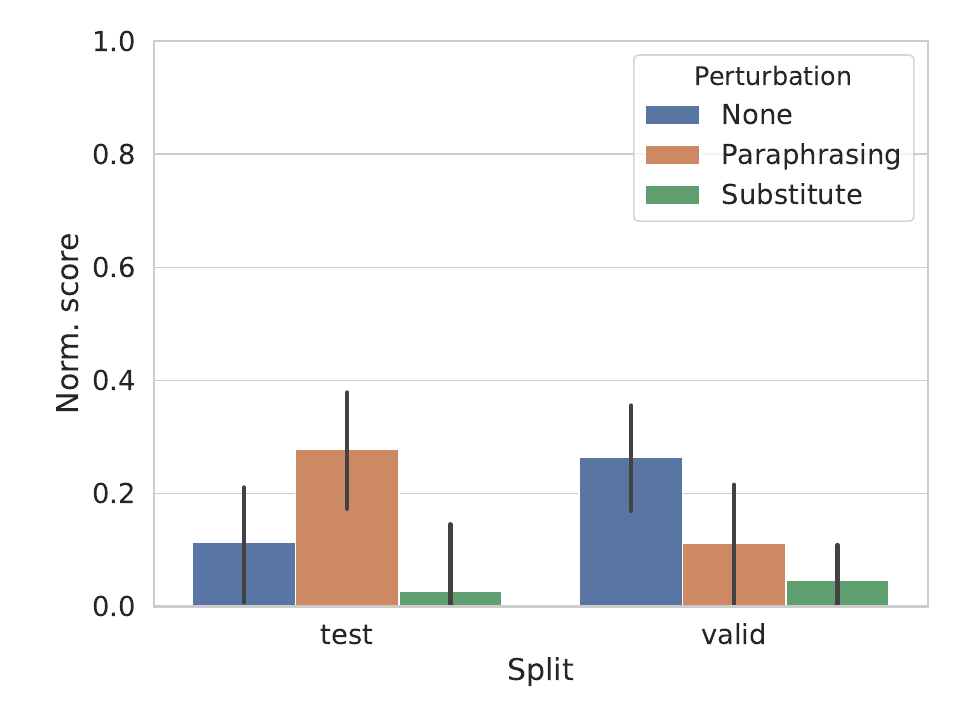}
        \caption{TWC Hard}
    \end{subfigure}

    \caption{Evaluation of an RoBERTa agent on original, paraphrased, and lexical substitution observations on (a) Easy, (b) Medium and (c) Hard games.}
    \label{fig:twc-perturbations-full}
\end{figure*}

Figure \ref{fig:twc-perturbations-full} shows the results for the perturbation experiments in TWC difficulties.

The result show how that a fixed LM model (RoBERTa) can maintain a relatively similar performance to the original observations when playing noisy versions of the game.

\subsection{Comparison of Fine-tuned/Fixed LMs on Jericho games}

\begin{figure*}[t]
    \centering
    \includegraphics[width=0.6\textwidth]{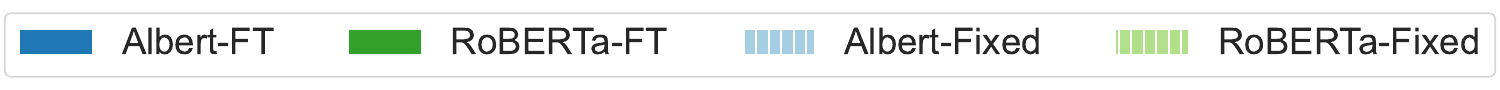}

    \begin{subfigure}{0.4\textwidth}
        \includegraphics[width=\textwidth]{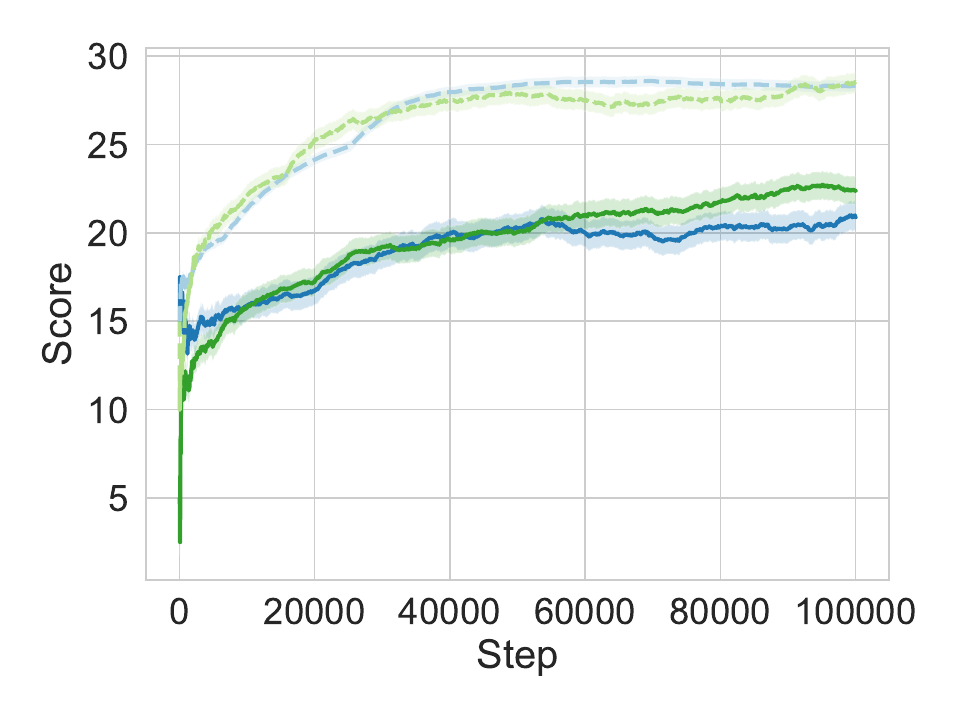}
        \caption{inhumane}
    \end{subfigure}~
    \begin{subfigure}{0.4\textwidth}
        \includegraphics[width=\textwidth]{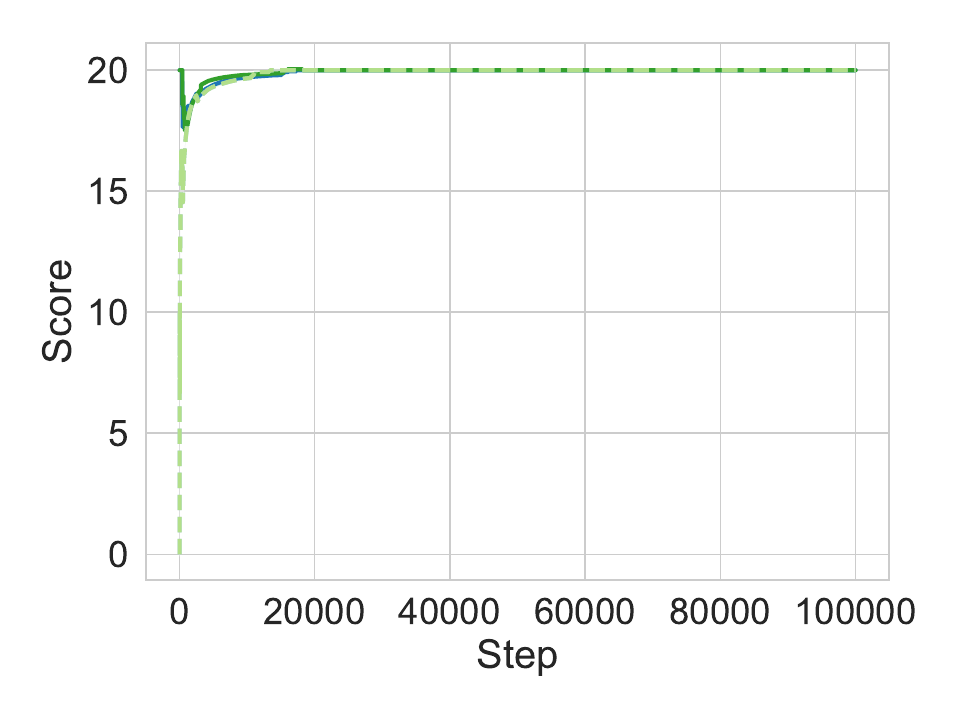}
        \caption{enchanter}
    \end{subfigure}
    
    \begin{subfigure}{0.4\textwidth}
        \includegraphics[width=\textwidth]{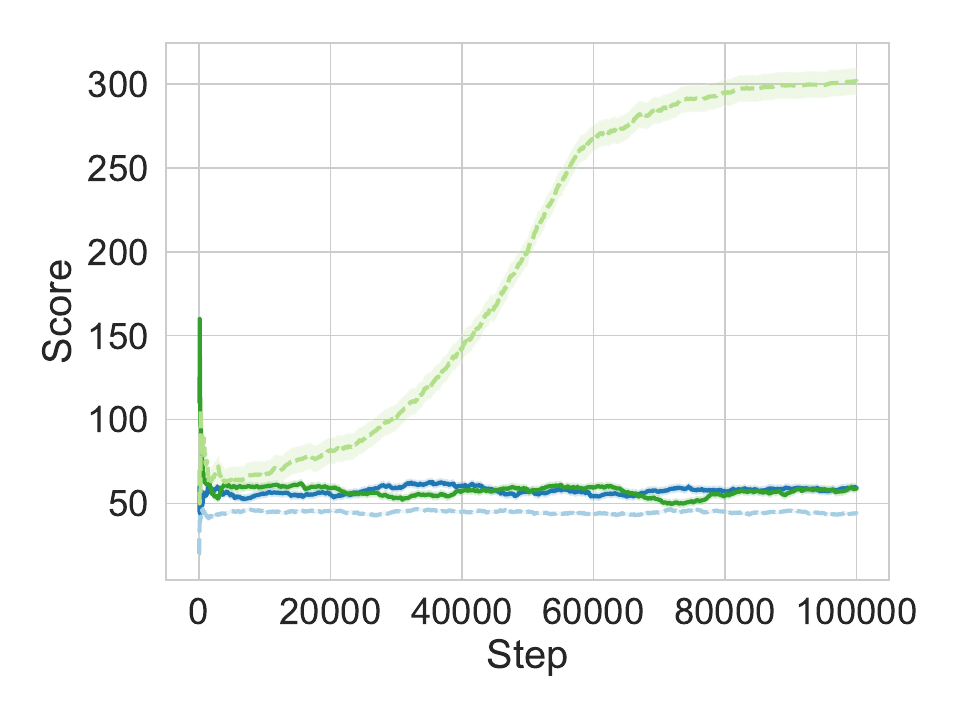}
        \caption{detective}
    \end{subfigure}~
    \begin{subfigure}{0.4\textwidth}
        \includegraphics[width=\textwidth]{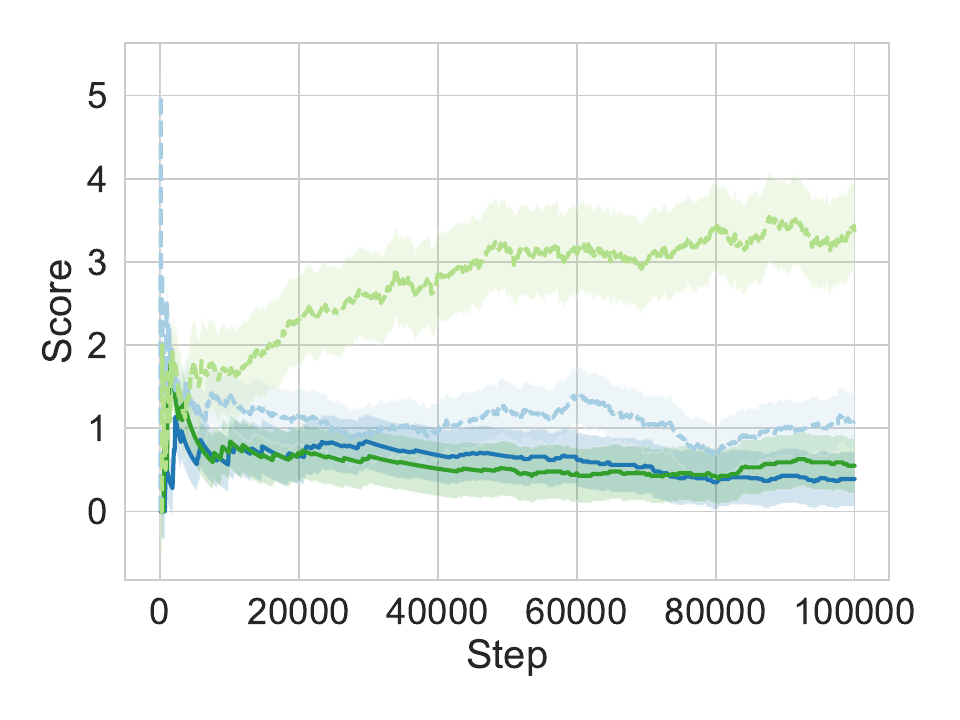}
        \caption{pentari}
    \end{subfigure}
    \caption{Comparison of fine-tuned/fixed LMs on various Jericho games.}
    \label{fig:ft-jericho-games}
\end{figure*}

Figure \ref{fig:ft-jericho-games} shows the fine-tuning/fixed LM comparison on additional games from the Jericho library: \texttt{detective}, \texttt{pentari}, \texttt{inhumane}, and \texttt{enchanter}.
The models show a consistent trend in which the fixed LMs outperform the fine-tuned models.

\subsection{Text perturbations}

This sections presents a description of the perturbations applied to the game texts.

A perturbation is a modification of an original piece of text in the game to produce an ``out-of-training'' example. Perturbations are applied to the observations, actions and inventories.

The types of perturbations are:

\begin{itemize}
    \item {Lexical substitution - we use WordNet synsets to find replacements for words in the text}
    \item {Paraphrasing - we use a sequence-to-sequence BART paraphraser to rephrase the original text}
\end{itemize}

\section {Reproducibility}

The code needed used to implement the methods described in this manuscript are submitted along with the supplementary material.
The code is anonymous and contains the instructions to set up the environments, download the game data, and train the agents.

\begin{table*}[ht]
    \centering
    \begin{tabular}{l|rr|rr|rr}
    & \multicolumn{2}{c}{\textbf{Easy}} & \multicolumn{2}{c}{\textbf{Medium}} & \multicolumn{2}{c}{\textbf{Hard}} \\ 
\textbf{Model}            & \multicolumn{1}{c}{Score} & \multicolumn{1}{c|}{Moves} & \multicolumn{1}{c}{Score} & \multicolumn{1}{c|}{Moves} & \multicolumn{1}{c}{Score} & \multicolumn{1}{c}{Moves} \\
    \toprule
    DRRN & $0.88 \pm 0.04$ & $24 \pm 2$ & $0.60 \pm 0.02$ & $44 \pm 1$ & $0.30 \pm 0.02$ & $50 \pm 0$ \\
    TPC & $0.89 \pm 0.06$ & $21 \pm 5$ & $0.62 \pm 0.03$ & $43 \pm 1$ & $0.32 \pm 0.04$ & $48 \pm 1$ \\
    KG-A2C & $0.86 \pm 0.06$ & $22 \pm 3$ & $0.62 \pm 0.03$ & $42 \pm 0$ & $0.32 \pm 0.00$ & $48 \pm 1$ \\
    BiKE & $0.94 \pm 0.00$ & $18 \pm 1$ & $0.64 \pm 0.02$ & $39 \pm 1$ & $0.34 \pm 0.00$ & $47 \pm 1$ \\
    BiKE + CBR & $0.95 \pm 0.04$ & $16 \pm 1$ & $0.67 \pm 0.03$ & $\textbf{35} \pm 1$ & $\textbf{0.42} \pm 0.04$ & $\textbf{45} \pm 1$ \\
    \midrule
    Hash & $0.31 \pm 0.07$ & $43 \pm 2$ & $0.58 \pm 0.06$ & $43 \pm 2$ & $0.22 \pm 0.03$ & $50 \pm 0$ \\
    Simple  & $0.83 \pm 0.08$ & $26 \pm 4$ & $0.58 \pm 0.08$ & $43 \pm 2$ & $0.35 \pm 0.05$ & $49 \pm 0$ \\
    Albert* & $0.96 \pm 0.02$ & $10 \pm 2$ & $0.66 \pm 0.05$ & $38 \pm 2$ & $ 0.41 \pm 0.05$ & $49 \pm 0$ \\
    MPNet* & $0.85 \pm 0.04$ & $19 \pm 3$ & $0.66 \pm 0.06$ & $38 \pm 2$ & $0.36 \pm 0.04$ & $49 \pm 0$ \\
    RoBERTa* & $0.94 \pm 0.03$ & $12 \pm 2$ & $\textbf{0.70} \pm 0.05$ & $38 \pm 2$ & $0.40 \pm 0.04$ & $49 \pm 0$ \\
    XLNet* & $1.00 \pm 0.00$ & $\textbf{6} \pm 1$ & $0.65 \pm 0.08$ & $36 \pm 3$ & $0.37 \pm 0.07$ & $48 \pm 1$ \\
    \end{tabular}
        \caption{Results for the in-distribution (valid) sets in TWC. (*) Indicates agents with fixed LM encoders.}
    \label{tab:twc-in-distro-full}
\end{table*}

\begin{table*}[h]
    \centering
    \begin{tabular}{l|rr|rr|rr}
    & \multicolumn{2}{c}{\textbf{Easy}} & \multicolumn{2}{c}{\textbf{Medium}} & \multicolumn{2}{c}{\textbf{Hard}} \\ 
\textbf{Model}            & \multicolumn{1}{c}{Score} & \multicolumn{1}{c|}{Moves} & \multicolumn{1}{c}{Score} & \multicolumn{1}{c|}{Moves} & \multicolumn{1}{c}{Score} & \multicolumn{1}{c}{Moves} \\
    \toprule
    DRRN & $0.78 \pm 0.02$ & $30 \pm 3$ & $0.55 \pm 0.01$ & $46 \pm 0$ & $0.20 \pm 0.02$ & $50 \pm 0$ \\
    TPC & $0.78 \pm 0.07$ & $28 \pm 4$ & $0.58 \pm 0.01$ & $45 \pm 2$ & $0.19 \pm 0.03$ & $50 \pm 0$ \\
    KG-A2C & $0.80 \pm 0.07$ & $28 \pm 4$ & $0.59 \pm 0.01$ & $43 \pm 3$ & $0.21 \pm 0.00$ & $50 \pm 0$ \\
    BiKE & $0.83 \pm 0.01$ & $26 \pm 2$ & $0.61 \pm 0.01$ & $41 \pm 2$ & $0.23 \pm 0.02$ & $50 \pm 0$ \\
    BiKE + CBR & $0.93 \pm 0.03$ & $17 \pm 1$ & $0.67 \pm 0.03$ & $35 \pm 1$ & $\textbf{0.40} \pm 0.03$ & $\textbf{46} \pm 1$ \\
    \midrule
    Simple & $0.50 \pm 0.12$ & $39 \pm 4$ & $0.43 \pm 0.07$ & $43 \pm 2$ & $0.26 \pm 0.04$ & $50 \pm 0$ \\
    Hash & $0.19 \pm 0.06$ & $44 \pm 2$ & $0.15 \pm 0.03$ & $50 \pm 0$ & $0.09 \pm 0.02$ & $50 \pm 0$ \\
    Albert* & $0.64 \pm 0.05$ & $33 \pm 3$ & $0.65 \pm 0.05$ & $\textbf{38} \pm 2$ & $0.16 \pm 0.02$ & $50 \pm 0$ \\
    MPNet* & $0.85 \pm 0.05$ & $23 \pm 2$ & $0.58 \pm 0.06$ & $42 \pm 2$ & $0.14 \pm 0.02$ & $50 \pm 0$ \\
    RoBERTa* & $0.90 \pm 0.04$ & $19 \pm 2$ & $0.53 \pm 0.06$ & $44 \pm 1$ & $0.19 \pm 0.03$ & $50 \pm 0$ \\
    XLNet* & $0.64 \pm 0.05$ & $30 \pm 3$ & $0.42 \pm 0.07$ & $47 \pm 1$ & $0.17 \pm 0.03$ & $50 \pm 0$ \\

    \end{tabular}
        \caption{Results for the out-of-distribution (test) sets in TWC. (*) Indicates agents with fixed LM encoders.}
    \label{tab:twc-out-of-distro-full}
\end{table*}
\end{document}